\documentclass{new_tlp}

\usepackage{times}
\usepackage{amsmath}
\usepackage{amsfonts}
\usepackage{amssymb}
\usepackage{graphicx}
\usepackage{mathabx}
\usepackage{listings}
\usepackage{color}
\usepackage{wrapfig}
\usepackage{url}
\usepackage{csquotes}
\newcommand{\cblu}{\color{blue}}

\def\sm{\hbox{\rm SM}}
\long\def\BOC#1\EOC{\message{(Commented text )}}
\long\def\BOCC#1\EOCC{\message{(Commented text )}}
\long\def\BOCCC#1\EOCCC{\message{(Commented text )}}
\long\def\BOCCCC#1\EOCCCC{\message{(Commented text )}}
\long\def\optional#1{\empty}

\long\def\NB#1{}

\def\ar{\leftarrow}
\def\beq{\begin{equation}}
\def\eeq#1{\label{#1}\end{equation}}
\def\bi{\begin{itemize}}
\def\ei{\end{itemize}}
\def\ii{\item}
\def\ba{\begin{array}}
\def\ea{\end{array}}
\def\mi#1{\hbox{\it #1\/}}

\def\sm{\hbox{\rm SM}}

\def\ar{\leftarrow}
\def\rar{\rightarrow}

\def\mvis{\!=\!}
\def\false{\hbox{\bf f}}
\def\true{\hbox{\bf t}}

\def\proof{\noindent{\bf Proof}.\hspace{3mm}}

\def\mvis{\!=\!}

\def\lpmln{\hbox{\rm LP}^{\rm{MLN}}}

\def\sneg{\sim\!\!}
\def\caused{\hbox{\bf caused}}
\def\causedab{\hbox{\bf caused\_ab}}
\def\iif{\hbox{\bf if}}
\def\after{\hbox{\bf after}}

\def\default{\hbox{\bf default}}

\def\caused{\hbox{\bf caused}}
\def\iif{\hbox{\bf if}}
\def\init{\hbox{\bf initially}}
\def\after{\hbox{\bf after}}

\def\default{\hbox{\bf default}}

\def\:{\!:\!}

\newtheorem{prop}{Proposition}
\newtheorem{thm}{Theorem}
\newtheorem{cor}{Corollary}

\newtheorem{example}{Example}

\submitted {}
\revised {}
\accepted{}

\begin{document}

\title{A Probabilistic Extension of Action Language ${\cal BC}$+} 

\author[Lee \& Wang]{Joohyung Lee and Yi Wang \\
School of Computing, Informatics and Decision Systems Engineering \\
Arizona State University, Tempe, USA \\
\email{\{joolee, ywang485\}@asu.edu}}

\maketitle

\begin{abstract}
We present a probabilistic extension of action language ${\cal BC}$+. Just like ${\cal BC}$+ is defined as a high-level notation of answer set programs for describing transition systems, the proposed language, which we call $p{\cal BC}$+, is defined as a high-level notation of $\lpmln$ programs---a probabilistic extension of answer set programs.
We show how probabilistic reasoning about transition systems, such as 
prediction, postdiction, and planning problems, as well as probabilistic diagnosis for dynamic domains, can be modeled in $p{\cal BC}$+ and computed using an implementation of $\lpmln$.

\noindent
(The paper is under consideration for acceptance in TPLP.)

\end{abstract}

\section{Introduction}

Action languages, such as ${\cal A}$ \cite{gel93a}, ${\cal B}$ \cite{gel98}, ${\cal C}$ \cite{giu98}, ${\cal C}$+ \cite{giu04}, and ${\cal BC}$ \cite{lee13answer}, are formalisms for describing actions and their effects. Many of these languages can be viewed as high-level notations of answer set programs structured to represent transition systems. 
The expressive possibility of action languages, such as indirect effects, triggered actions, and additive fluents, has been one of the main research topics.
Most of such extensions are logic-oriented, and less attention has been paid to probabilistic reasoning, with a few exceptions such as \cite{baral02reasoning,eiter03probabilistic}, let alone automating such probabilistic reasoning and learning parameters of an action description.

Action language ${\cal BC}$+ \cite{babb15action1}, one of the most recent additions to the family of action languages, is no exception. While the language is highly expressive to embed other action languages, such as ${\cal C}$+ \cite{giu04} and ${\cal BC}$ \cite{lee13action}, it does not have a natural way to express the probabilities of histories (i.e., a sequence of transitions). 

In this paper, we present a probabilistic extension of ${\cal BC}$+, which we call $p{\cal BC}$+. Just like ${\cal BC}$+ is defined as a high-level notation of answer set programs for describing transition systems, $p{\cal BC}$+ is defined as a high-level notation of $\lpmln$ programs---a probabilistic extension of answer set programs.
Language $p{\cal BC}$+ inherits expressive logical modeling capabilities of ${\cal BC}$+ but also allows us to assign a probability to a sequence of transitions so that we may distinguish more probable histories. 

We show how probabilistic reasoning about transition systems, such as prediction, postdiction, and planning problems, can be modeled in $p{\cal BC}$+ and computed using an implementation of $\lpmln$.
Further, we show that it can be used for probabilistic abductive reasoning about dynamic domains, where the likelihood of the abductive explanation is derived from the parameters manually specified or automatically learned from the data. 
%
%
%
%
%
%

\BOCC
The computation of the language can be carried out based on the recent progress with $\lpmln$. The inference engine {\sc lpmln2asp} \cite{lee17computing} translates $\lpmln$ programs into the input language of answer set solver {\sc clingo}, and using weak constraints and stable model enumeration, it can compute most probable stable models as well as exact conditional and marginal probabilities.
\EOCC

The paper is organized as follows. Section~\ref{sec:prelim} reviews language $\lpmln$ and multi-valued probabilistic programs that are defined in terms of $\lpmln$. 
Section~\ref{sec:pbc} presents language $p{\cal BC}$+, and Section~\ref{sec:inference} shows how to use $p{\cal BC}$+ and system {\sc lpmln2asp} \cite{lee17computing} to perform probabilistic reasoning about transition systems, such as prediction, postdiction, and planning.  Section~\ref{sec:diagnosis} extends $p{\cal BC}$+ to handle probabilistic diagnosis.




\section{Preliminaries}\label{sec:prelim}

\subsection{Review: Language $\lpmln$}

{
We review the definition of $\lpmln$ from~\cite{lee16weighted}, limited to the propositional case.
An $\lpmln$ program is a finite set of weighted rules $w: R$ where $R$ is a propositional formula, $w$ is a real number (in which case, the weighted rule is called {\em soft}) or $\alpha$ for denoting the infinite weight (in which case, the weighted rule is called {\em hard}). 
}

For any $\lpmln$ program $\Pi$ and any interpretation~$I$, 
$\overline{\Pi}$ denotes the usual (unweighted) ASP program obtained from $\Pi$ by dropping the weights, and
${\Pi}_I$ denotes the set of $w: R$ in $\Pi$ such that $I\models R$, 
and $\sm[\Pi]$ denotes the set $\{I \mid \text{$I$ is a stable model of $\overline{\Pi_I}$}\}$.
The {\em unnormalized weight} of an interpretation $I$ under $\Pi$ is defined as 
\[
 W_\Pi(I) =
\begin{cases}
  exp\Bigg(\sum\limits_{w:R\;\in\; {\Pi}_I} w\Bigg) & 
      \text{if $I\in\sm[\Pi]$}; \\
  0 & \text{otherwise}. 
\end{cases}
\]
The {\em normalized weight} (a.k.a. {\em probability}) of an interpretation $I$ under~$\Pi$ is defined as  \[ 
\small 
  P_\Pi(I)  = 
  \lim\limits_{\alpha\to\infty} \frac{W_\Pi(I)}{\sum\limits_{J\in {\rm SM}[\Pi]}{W_\Pi(J)}}. 
\] 
Interpretation $I$ is called a {\sl (probabilistic) stable model} of $\Pi$ if $P_\Pi(I)\ne 0$. The most probable stable models of $\Pi$ are the stable models with the highest probability.

\NB{
{\cred [[Mention {\sc lpmln2asp} and lpmln learning?]]}}

\subsection{Review: Multi-Valued Probabilistic Programs} \label{ssec:mvpp}

Multi-valued probabilistic programs \cite{lee16weighted} are a simple fragment of $\lpmln$ that allows us to represent probability more naturally. 

We assume that the propositional signature $\sigma$ is constructed from ``constants'' and their ``values.'' 
A {\em constant} $c$ is a symbol that is associated with a finite set $\mi{Dom}(c)$, called the {\em domain}. 
The signature $\sigma$ is constructed from a finite set of constants, consisting of atoms $c\!=\!v$~\footnote{%
Note that here ``='' is just a part of the symbol for propositional atoms, and is not  equality in first-order logic. }
for every constant $c$ and every element $v$ in $\mi{Dom}(c)$.
If the domain of~$c$ is $\{\false,\true\}$ then we say that~$c$ is {\em Boolean}, and abbreviate $c\mvis\true$ as $c$ and $c\mvis\false$ as~$\sneg c$. 

We assume that constants are divided into {\em probabilistic} constants and {\em non-probabilistic} constants.
A multi-valued probabilistic program ${\bf \Pi}$ is a tuple $\langle \mi{PF}, \Pi \rangle$, where
\begin{itemize}
\item $\mi{PF}$ contains \emph{probabilistic constant declarations} of the following form:
\begin{equation}\label{eq:probabilistic-constant-declaration}
p_1::\ c\mvis v_1\mid\dots\mid p_n::\ c\mvis v_n
\end{equation}
one for each probabilistic constant $c$, where $\{v_1,\dots, v_n\}=\mi{Dom}(c)$, $v_i\ne v_j$, $0\leq p_1,\dots,p_n\leq1$ and $\sum_{i=1}^{n}p_i=1$. We use $M_{\bf \Pi}(c=v_i)$ to denote $p_i$.
In other words, $\mi{PF}$ describes the probability distribution over each ``random variable''~$c$. 

\item $\Pi$ is a set of rules of the form $\mi{Head}\ar\mi{Body}$ (identified with formula $\mi{Body}\rar\mi{Head}$ such that  $\mi{Head}$ and $\mi{Body}$ do not contain implications, and $\mi{Head}$ contains no probabilistic constants.
\end{itemize}

The semantics of such a program ${\bf \Pi}$ is defined as a shorthand for $\lpmln$ program $T({\bf \Pi})$ of the same signature as follows.
\begin{itemize}
\item For each probabilistic constant declaration (\ref{eq:probabilistic-constant-declaration}), $T({\bf \Pi})$ contains, 
for each $i=1,\dots, n$,
(i) $ln(p_i):  c\mvis v_i$  if $0<p_i<1$; 
(ii) $\alpha:\ c\mvis v_i$ if $p_i=1$;
(iii) $\alpha:\ \bot\ar c\mvis v_i$ if $p_i=0$.

\item  For each rule $\mi{Head}\ar\mi{Body}$ in $\Pi$, $T({\bf \Pi})$ contains
$
\alpha:\ \  \mi{Head}\ar\mi{Body}. 
$

\ii For each constant $c$, $T({\bf \Pi})$ contains the uniqueness of value constraints
\beq
\ba {rl}
   \alpha: & \bot \ar c\mvis v_1\land c=v_2 
\ea 
\eeq{uc}
for all $v_1,v_2 \in\mi{Dom}(c)$ such that $v_1\ne v_2$, and the existence of value constraint
\beq
\ba {rl}
  \alpha: & \bot \ar \neg \bigvee\limits_{v \in {Dom}(c)} c\mvis v\ .
\ea 
\eeq{ec}


\NB{{\cred [[Introduce the notion of total choice?]]}}
\end{itemize}

In the presence of the constraints \eqref{uc} and \eqref{ec}, assuming $T({\bf \Pi})$ has at least one (probabilistic) stable model that satisfies all the hard rules, a (probabilistic) stable model $I$ satisfies $c=v$ for exactly one value $v$, so we may identify $I$ with the value assignment that assigns $v$ to $c$.

\section{Probabilistic ${\cal BC}$+}\label{sec:pbc}

\subsection{Syntax}

We assume a propositional signature~$\sigma$ as defined in Section~\ref{ssec:mvpp}.
We further assume that the signature of an action description is divided into four groups: {\em fluent constants}, {\em action constants},  {\em pf (probability fact) constants}, and {\em  initpf (initial probability fact) constants}. Fluent constants are further divided into {\em regular} and {\em statically determined}. The domain of every action constant is Boolean. 
A {\em fluent formula} is a formula such that all constants occurring in it are fluent constants. 

The following definition of $p\cal{BC}$+ is based on the definition of ${\cal BC}$+ language from \cite{babb15action1}.

A {\em static law} is an expression of the form
\begin{equation}\label{eq:static-law}
\caused\ F\ \iif\ G
\end{equation}
where $F$ and $G$ are fluent formulas.


A {\em fluent dynamic law} is an expression of the form
\begin{equation}\label{eq:fluent-dynamic-law}
\caused\ F\ \iif\ G\ \after\ H
\end{equation}
where $F$ and $G$ are fluent formulas and $H$ is a formula, provided that $F$ does not contain statically determined constants and $H$ does not contain initpf constants.

A {\em pf constant declaration} is an expression of the form
\begin{equation}\label{eq:pf-declare-no-time}
   \caused\ \mi{c}=\{v_1:p_1, \dots, v_n:p_n\}
\end{equation}
where $\mi{c}$ is a pf constant with domain $\{v_1, \dots, v_n\}$, $0<p_i<1$ for each $i\in\{1, \dots, n\}$\footnote{We require $0<p_i<1$ for each $i\in\{1, \dots, n\}$ for the sake of simplicity. On the other hand, if $p_i=0$ or $p_i=1$ for some $i$, that means either $v_i$ can be removed from the domain of $c$ or there is not really a need to introduce $c$ as a pf constant. So this assumption does not really sacrifice expressivity.}, and $p_1+\cdots+p_n=1$. In other words, \eqref{eq:pf-declare-no-time} describes the probability distribution of $c$.

An {\em initpf constant declaration} is an expression of the form (\ref{eq:pf-declare-no-time}) where $c$ is an initpf constant. 

An {\em initial static law} is an expression of the form
\begin{equation}\label{eq:init-static-law}
\init\ F\ \iif\ G
\end{equation}
where $F$ is a fluent constant and $G$ is a formula that contains neither action constants nor pf constants. 

A {\em causal law} is a static law, a fluent dynamic law, a pf constant  declaration, an initpf constant declaration, or an initial static law. An {\em action description} is a finite set of causal laws.


We use $\sigma^{fl}$ to denote the set of fluent constants, $\sigma^{act}$ to denote the set of action constants, $\sigma^{pf}$ to denote the set of pf constants, and $\sigma^{initpf}$ to denote the set of initpf constants. For any signature $\sigma^\prime$ and any $i\in\{0, \dots, m\}$, we use $i:\sigma^\prime$ to denote the set
$\{i:a \mid a\in\sigma^\prime\}$.


By $i:F$ we denote the result of inserting $i:$ in front of every occurrence of every constant in formula $F$. This notation is straightforwardly extended when $F$ is a set of formulas.

\begin{example}\label{eg:simple-pt}
The following is an action description in $p{\cal BC}$+ for the transition  system shown in Figure~\ref{fig:psd}, $P$ is a Boolean regular fluent constant, and $A$ is an action constant. Action $A$ toggles the value of $P$ with probability $0.8$. Initially, $P$ is true with probability $0.6$ and false with probability $0.4$. We call this action description $\mi{PSD}$. ($x$ is a schematic variable that ranges over $\{\true, \false\}$.)

\begin{minipage}[c]{0.4\textwidth}
\[
\ba l
\caused\ P\ \iif\ \top\ \after\ \sneg P\wedge A\wedge \mi{Pf},\\
\caused\ \sneg P\ \iif\ \top\ \after\ P\wedge A\wedge \mi{Pf},\\
\caused \ \left\{P\right\}^{\rm ch} \iif\ \top\ \after\ P, \\
\caused \ \left\{\sneg P\right\}^{\rm ch} \iif\ \top\ \after\ \sneg P, 
\ea
\]
\end{minipage}
\begin{minipage}[c]{0.5\textwidth} 
\[
\ba l
\caused\ \mi{Pf}=\{\true: 0.8, \false: 0.2\},\\
\caused\ \mi{InitP}=\{\true: 0.6, \false: 0.4\},\\
\init\ P=x\ \iif\ \mi{InitP}=x. \\ \\
\ea
\]
\end{minipage}

\smallskip\noindent
($\{P\}^{\rm ch}$ is a choice formula standing for $P\lor \neg P$.)

\begin{figure}
  \centering
  \includegraphics[width=0.65\textwidth]{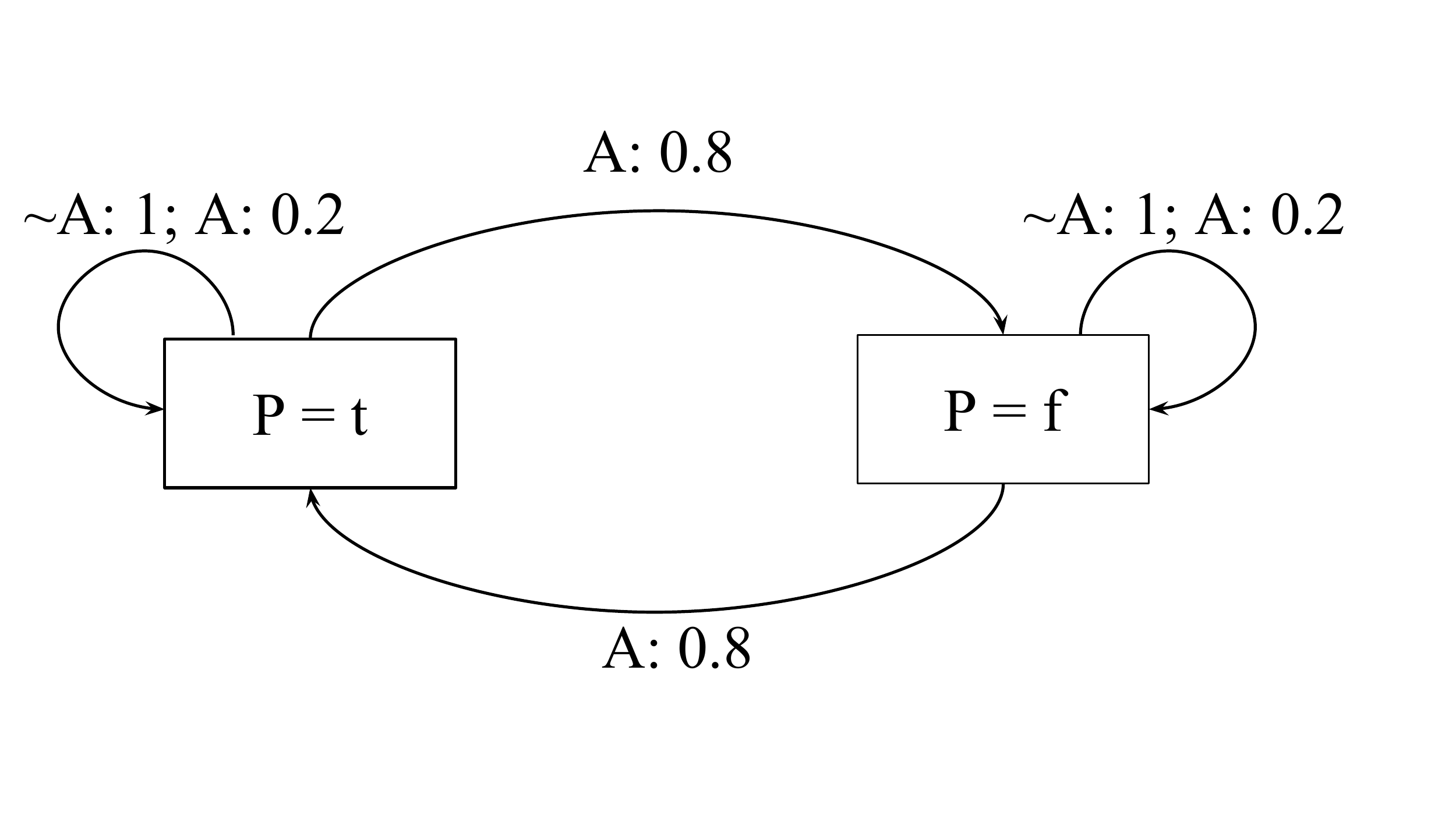}
  \vspace{-0.7cm}
  \caption{A transition system with probabilistic transitions}
  \label{fig:psd}
\end{figure} 

\end{example}

\subsection{Semantics}

Given a non-negative integer $m$ denoting the maximum length of histories, the semantics of an action description $D$ in $p{\cal BC}$+ is defined by a reduction to multi-valued probabilistic program $Tr(D, m)$, which is the union of two subprograms $D_m$ and $D_{init}$ as defined below. 

For an action description $D$ of a signature $\sigma$, we define a sequence of multi-valued probabilistic program $D_0, D_1, \dots,$ so that the stable models of $D_m$ can be identified with the paths in the transition system described by $D$.
The signature $\sigma_m$ of $D_m$ consists of atoms of the form $i:c=v$ such that
\begin{itemize}
\item for each fluent constant $c$ of $D$, $i\in\{0, \dots, m\}$ and $v\in Dom(c)$,
\item for each action constant or pf constant $c$ of $D$, $i\in\{0, \dots, m-1\}$ and $v\in Dom(c)$.
\end{itemize}

For $x\in \{act, fl, pf\}$, we use $\sigma^{x}_m$ to denote the subset of $\sigma_m$
\[
\{i:c=v \mid \text{$i:c=v\in \sigma_m$ and $c\in\sigma^{x}$}\}.
\]
For $i\in\{0, \dots, m\}$, we use $i:\sigma^{x}$ to denote the subset of $\sigma_m^{x}$
\[
\{i:c=v\mid i:c=v\in \sigma_m^{x}\}.
\]

We define $D_m$ to be the multi-valued probabilistic program  $\langle PF, \Pi\rangle$, where $\Pi$ is the conjunction of
\begin{equation}\label{eq:static-law-asp}
i:F \leftarrow i:G
\end{equation}
for every static law (\ref{eq:static-law}) in $D$ and every $i\in\{0, \dots, m\}$,
\begin{equation}\label{eq:fluent-dynamic-law-asp}
i\!+\!1:F\leftarrow (i\!+\!1:G)\wedge(i:H)
\end{equation}
for every fluent dynamic law (\ref{eq:fluent-dynamic-law}) in $D$ and every $i\in\{0, \dots, m-1\}$,
\begin{equation}\label{eq:init-fluent-choice}
\{0\!:\!c=v\}^{\rm ch}
\end{equation}
for every regular fluent constant $c$ and every $v\in Dom(c)$,
\begin{equation}\label{eq:action-choice}
\{i:c=\true\}^{\rm ch}, \ \ \ \ 
\{i:c=\false\}^{\rm ch}
\end{equation}
for every action constant $c$,
and $PF$ consists of 
\begin{equation}\label{eq:prod-declaration-mvpp}
   p_1::\ i:pf=v_1 \mid \dots \mid p_n::\ i:pf=v_{n}
\end{equation}
($i=0,\dots,m-1$) for each pf constant declaration \eqref{eq:pf-declare-no-time} in $D$ that describes the probability distribution of $\mi{pf}$.

Also, we define the program $D_{init}$, whose signature is $0\!:\!\sigma^{initpf}\cup 0\!:\!\sigma^{fl}$.
$D_{init}$ is the multi-valued probabilistic program
\[
D_{init} = \langle PF^{init}, \Pi^{init}\rangle
\]
where $\Pi^{init}$ consists of the rule
\[
\bot\leftarrow \neg(0\!:\!F)\land 0\!:\!G
\]
for each initial static law (\ref{eq:init-static-law}),
and $PF^{init}$ consists of 
\[
p_1::\ 0\!:\!pf=v_1\ \ \mid\ \  \dots\ \  \mid\ \  p_n::\ 0\!:\!pf=v_n
\]
for each initpf constant declaration (\ref{eq:pf-declare-no-time}).

We define $Tr(D, m)$ to be the union of the two multi-valued probabilistic program \\
$
\langle  PF\cup PF^{init}, \Pi\cup\Pi^{init} \rangle.
$

\begin{example}
For the action description $\mi{PSD}$ in Example \ref{eg:simple-pt}, $\mi{PSD}_{init}$ is the following multi-valued probabilistic program ($x\in\{\true, \false\}$):
\[
\ba l
  0.6 ::\ 0\!:\!\mi{InitP} \ \mid\ 0.4 ::\ 0\!:\sneg \mi{InitP}\\
  \bot \leftarrow \neg (0\!:\!P\mvis x)\land 0:\mi{InitP}\mvis x.
\ea 
\]
and $\mi{PSD}_m$ is the following multi-valued probabilistic program ($i$ is a schematic variable that ranges over $\{1, \dots, m-1\}$):

\begin{minipage}[c]{0.4\textwidth}
\[
\ba l
0.8::\ i:\mi{Pf}\ \mid\ 0.2::\ i:\sneg \mi{Pf}\\
i\!+\!1:P \leftarrow i:\sneg P\wedge i:A\wedge i:\mi{Pf}\\
i\!+\!1:\sneg P \leftarrow i:P\wedge i:A\wedge i:\mi{Pf} \\ 
\{i\!+\!1:P\}^{\rm ch}\leftarrow i:P\\
\{i\!+\!1:\sneg P\}^{\rm ch}\leftarrow i:\sneg P
\ea 
\]
\end{minipage}
\begin{minipage}[c]{0.5\textwidth}
\[
\ba l
\{i:A\}^{\rm ch}\\
\{i:\sneg A\}^{\rm ch}\\
\{0\!:\!P\}^{\rm ch}\\ 
\{0\!:\!\sneg P\}^{\rm ch}
\ea 
\]
\end{minipage}
\end{example}

For any $\lpmln$ program $\Pi$ of signature $\sigma$ and a value assignment $I$ to a subset $\sigma'$ of $\sigma$, we say $I$ is a {\em residual (probabilistic) stable model} of $\Pi$ if there exists a value assignment $J$ to $\sigma\setminus \sigma'$ such that $I\cup J$ is a (probabilistic) stable model of $\Pi$.

For any value assignment $I$ to constants in $\sigma$, by $i\!:\!I$ we denote the value assignment to constants in $i\!:\!\sigma$ so that $i\!:\!I\models (i\!:\!c)=v$ iff $I\models c=v$.

We define a {\em state} as an interpretation $I^{fl}$ of $\sigma^{fl}$ such that $0\!:\!I^{fl}$ is a residual (probabilistic) stable model of $D_0$. A {\em transition} of $D$ is a triple $\langle s, e, s^\prime\rangle$  where $s$ and $s^\prime$ are interpretations of $\sigma^{fl}$ and $e$ is an interpretation of $\sigma^{act}$ such that $0\!:\!s \cup 0\!:\!e \cup 1:s^\prime$ is a residual stable model of $D_1$. A {\em pf-transition} of $D$ is a pair $(\langle s, e, s^\prime\rangle, pf)$, where $pf$ is a value assignment to $\sigma^{pf}$ such that $0\!:\!s\cup 0\!:\!e \cup 1:s^\prime \cup 0\!:\!pf$ is a stable model of $D_1$.


A {\em probabilistic transition system} $T(D)$ represented by a probabilistic action description $D$ is a labeled directed graph such that the vertices are the states of $D$, and the edges are obtained from the transitions of $D$: for every transition $\langle s, e, s^\prime\rangle$  of $D$, an edge labeled $e: p$ goes from $s$ to $s^\prime$, where $p=Pr_{D_m}(1\!:\!s^\prime \mid 0\!:\!s, 0\!:\!e)$. The number $p$ is called the {\em transition probability} of $\langle s, e ,s^\prime\rangle$ .

The soundness of the definition of a probabilistic transition system relies on the following proposition. 
\begin{prop}\label{prop:state-in-transition}
For any transition $\langle s, e, s^\prime \rangle$, $s$ and $s^\prime$ are states.
\end{prop}

We make the following simplifying assumptions on action descriptions:

\begin{enumerate}
\item {\bf No Concurrency}: For all transitions $\langle s, e, s'\rangle$, we have $e(a)=t$ for at most one $a\in \sigma^{act}$;
\item {{\bf Nondeterministic Transitions are Controlled by pf constants}:} For any state $s$, any value assignment $e$ of $\sigma^{act}$ such that at most one action is true, and any value assignment $pf$ of $\sigma^{pf}$, there exists exactly one state $s^\prime$ such that $(\langle s, e, s^\prime\rangle, pf)$ is a pf-transition;
\item {\bf Nondeterminism on Initial States are Controlled by Initpf constants}: Given any assignment $pf_{init}$ of $\sigma^{initpf}$, there exists exactly one assignment $fl$ of $\sigma^{fl}$ such that $0\!:\!pf_{init}\cup 0\!:\!fl$ is a stable model of $D_{init}\cup D_0$.

\end{enumerate}

For any state $s$, any value assignment $e$ of $\sigma^{act}$ such that at most one action is true, and any value assignment $pf$ of $\sigma^{pf}$, we use $\phi(s, e, pf)$ to denote the state $s'$ such that $(\langle s, a, s^\prime\rangle, pf)$ is a pf-transition (According to Assumption 2, such $s^\prime$ must be unique). For any interpretation $I$, $i\in \{0, \dots, m\}$ and any subset $\sigma^\prime$ of $\sigma$, we use $I|_{i:\sigma^\prime}$ to denote the value assignment of $I$ to atoms in $i:\sigma^\prime$. Given any value assignment $TC$ of $0\!:\!\sigma^{initpf}\cup \sigma^{pf}_m$and a value assignment $A$ of $\sigma_m^{act}$, we construct an interpretation $I_{TC\cup A}$ of $Tr(D, m)$ that satisfies $TC \cup A$ as follows:
\begin{itemize}
\item  For all atoms $p$ in $\sigma^{pf}_m\cup 0\!:\!\sigma^{initpf}$, 
           we have $I_{TC\cup A}(p) = TC(p)$;
\item  For all atoms $p$ in $\sigma_m^{act}$, we have $I_{TC\cup A}(p) = A(p)$;
\item $(I_{TC\cup A})|_{0:\sigma^{fl}}$ is the assignment such that $(I_{TC\cup A})|_{0:\sigma^{fl}\cup 0:\sigma^{initpf}}$ is a stable model of $D_{init}\cup D_0$.
\item For each $i\in \{1, \dots, m\}$, $$(I_{TC\cup A})|_{i:\sigma^{fl}} = \phi((I_{TC\cup A})|_{(i-1):\sigma^{fl}}, (I_{TC\cup A})|_{(i-1):\sigma^{act}}, (I_{TC\cup A})|_{(i-1):\sigma^{pf}}).$$
\end{itemize}
By Assumptions 2 and 3, the above construction produces a unique interpretation. 

It can be seen that in the multi-valued probabilistic program $Tr(D, m)$ translated from $D$, the probabilistic constants  are $0\!:\!\sigma^{initpf}\cup \sigma^{pf}_m$. We thus call the value assignment of an interpretation $I$ on $0\!:\!\sigma^{initpf}\cup \sigma^{pf}_m$ the {\em total choice} of $I$. The following theorem asserts that the probability of a stable model under $Tr(D, m)$ can be computed by simply dividing the probability of the total choice associated with the stable model by the number of choice of actions.

\begin{thm}\label{thm:path-probability}
For any value assignment $TC$ of $ 0\!:\!\sigma^{initpf}\cup\sigma^{pf}_m$ and any value assignment $A$ of $\sigma_m^{act}$, there exists exactly one stable model $I_{TC\cup A}$ of $Tr(D, m)$ that satisfies $TC\cup A$, and the probability of $I_{TC\cup A}$ is
\[
Pr_{Tr(D, m)}(I_{TC\cup A}) = \frac{\underset{c=v\in TC}{\prod}M(c=v)}{(|\sigma^{act}| + 1)^{m}}.
\]
\end{thm}

The following theorem tells us that the conditional probability of transiting from a state $s$ to another state $s^\prime$ with action $e$ remains the same for all timesteps, i.e., the conditional probability of $i\!+\!1\!:\!s^\prime$ given $i:s$ and $i:e$ correctly represents the transition probability from $s$ to $s^\prime$ via $e$ in the transition system.

\begin{thm}\label{thm:transition-probability}
For any state $s$ and $s^\prime$, and any interpretation $e$ of $\sigma^{act}$, we have
\[
{Pr_{Tr(D, m)}(i\!+\!1\!:\!s^\prime\mid i:s, i:e) = Pr_{Tr(D, m)}(j\!+\!1\!:\!s^\prime\mid j:s, j:e)}
\]
for any $i, j\in\{0, \dots, m-1\}$ such that $Pr_{Tr(D, m)}(i:s)> 0$ and $Pr_{Tr(D, m)}(j:s)> 0$.
\end{thm}

For every subset $X_m$ of $\sigma_m\setminus\sigma^{pf}_m$, let $X^i(i < m)$ be the triple consisting of
\begin{itemize}
\item the set consisting of atoms $A$ such that $i:A$ belongs to $X_m$ and $A\in \sigma^{fl}$;
\item the set consisting of atoms $A$ such that $i:A$ belongs to $X_m$ and $A\in \sigma^{act}$;
\item the set consisting of atoms $A$ such that $i\!+\!1\!:\!A$ belongs to $X_m$ and $A\in \sigma^{fl}$.
\end{itemize}
Let $p(X^i)$ be the transition probability of $X^i$, $s_0$ is the interpretation of $\sigma^{fl}_0$ defined by $X^0$, and $e_i$ be the interpretations of $i:\sigma^{act}$ defined by $X^{i}$.

Since the transition probability remains the same, the probability of a path given a sequence of actions can be computed from the probabilities of transitions.

\begin{cor}\label{thm:reduce2transition}
For every $m\geq 1$, $X_m$ is a residual (probabilistic) stable model of $Tr(D, m)$ iff $X^0, \dots, X^{m-1}$ are transitions of $D$ and $0\!:\!s_0$ is a residual stable model of $D_{init}$. Furthermore, 
\[
Pr_{Tr(D, m)}(X_m\mid 0\!:\!e_0, \dots, m-1\!:\!e_{m-1}) = p(X^0)\times\dots\times p(X^m)\times Pr_{Tr(D, m)}(0\!:\!s_0).
\]
\end{cor}

\begin{example}
Consider the simple transition system with probabilistic effects in Example \ref{eg:simple-pt}. Suppose $a$ is executed twice. What is the probability that $P$ remains true the whole time? Using Corollary \ref{thm:reduce2transition} this can be computed as follows:
\[
\ba l
\small 
Pr(2:P=\true, 1\!:\!P=\true, 0\!:\!P=\true\mid 0\!:\!A=\true, 1\!:\!A=\true)\\
 = p(\langle P=\true, A=\true, P=\true\rangle)\cdot p(\langle P=\true, A=\true, P=\true\rangle)\cdot Pr_{Tr(D, m)}(0\!:\!P=\true)\\
 =\  0.2\times 0.2 \times 0.6 = 0.024.
\ea 
\]
\end{example}

\section{$p{\cal BC}$+ Action Descriptions and Probabilistic Reasoning} \label{sec:inference}


In this section, we illustrate how the probabilistic extension of the reasoning tasks discussed in \cite{giu04}, i.e., prediction, postdiction and planning, can be represented in $p\cal{BC}$+ and automatically computed using {\sc lpmln2asp} \cite{lee17computing}. Consider the following probabilistic variation of the well-known Yale Shooting Problem: There are two (slightly deaf) turkeys: a fat turkey and a slim turkey. Shooting at a turkey may fail to kill the turkey. Normally, shooting at the slim turkey has $0.6$ chance to kill it, and shooting at the fat turkey has $0.9$ chance. However, when a turkey is dead, the other turkey becomes alert, which decreases the success probability of shooting. For the slim turkey, the probability drops to $0.3$, whereas for the fat turkey, the probability drops to $0.7$.

The example can be modeled in $p\cal{BC}$+ as follows. First, we declare  the constants:

\vspace{0.2cm}
\hrule
\begin{tabbing}
Notation:  $t$ range over $\{\mi{SlimTurkey}, \mi{FatTurkey}\}$. \\
Regular fluent constants:          \hskip 4cm  \=Domains:\\
$\;\;\;$ $\mi{Alive}(t)$,  $\;$ $\mi{Loaded}$                 \>$\;\;\;$ Boolean\\ 
Statically determined fluent constants:     \hskip 2cm  \=Domains:\\
$\;\;\;$ $\mi{Alert}(t)$                 \>$\;\;\;$ Boolean\\ 
Action constants:                          \>Domains:\\
$\;\;\;$ $\mi{Load}$ , $\;$ $\mi{Fire}(t)$  \>$\;\;\;$ Boolean\\ 
Pf constants:                          \>Domains:\\
$\;\;\;$ $\mi{\mi{Pf\_Killed}}(t)$, $\;$ $\mi{\mi{Pf\_Killed\_Alert}}(t)$                    \>$\;\;\;$ Boolean \\
InitPf constants: \\
$\;\;\;$ $\mi{Init\_Alive}(t)$,  $\;$ $\mi{Init\_Loaded}$                 \>$\;\;\;$ Boolean
\end{tabbing}
\hrule
\vspace{0.2cm}

Next, we state the causal laws.
The effect of loading the gun is described by 
\begin{tabbing}
\ \ \ $\caused\ \mi{Loaded}\ \iif\ \top\ \after\ \mi{Load}$.
\end{tabbing}
To describe the effect of shooting at a turkey, we declare the following probability distributions on the result of shooting at each turkey when it is not alert and when it is alert:
\begin{tabbing}
\ \ \ $\caused\ \mi{Pf\_Killed}(\mi{SlimTurkey})=\{\true: 0.6, \false: 0.4\}$,  \\ 
\ \ \ $\caused\ \mi{Pf\_Killed}(\mi{FatTurkey})=\{\true: 0.9, \false: 0.1\}$, \\
\ \ \ $\caused\ \mi{Pf\_Killed\_Alert}(\mi{SlimTurkey})=\{\true: 0.3, \false: 0.7\}$,\\
\ \ \ $\caused\ \mi{Pf\_Killed\_Alert}(\mi{FatTurkey})=\{\true: 0.7, \false: 0.3\}$.
\end{tabbing}
The effect of shooting at a turkey is described as
\begin{tabbing}
\ \ \ $\caused\ \sneg \mi{Alive}(t)\ \iif\ \top\ \after\ \mi{Loaded}\wedge \mi{Fire}(t)\wedge \sneg \mi{Alert}(t)\wedge \mi{Pf\_Killed}(t)$,\\
\ \ \ $\caused\ \sneg \mi{Alive}(t)\ \iif\ \top\ \after\ \mi{Loaded}\wedge \mi{Fire}(t)\wedge \mi{Alert}(t)\wedge \mi{Pf\_Killed\_Alert}(t)$,\\
\ \ \ $\caused\ \sneg \mi{Loaded}\ \iif\ \top\ \after\ \mi{Fire}(t)$.
\end{tabbing}
A dead turkey causes the other turkey to be alert:
\begin{tabbing}
\ \ \ $\default\ \sneg \mi{Alert}(t)$, \\ 
\ \ \ $\caused\ \mi{Alert}(t_1)\ \iif\ \sneg \mi{Alive}(t_2)\wedge \mi{Alive}(t_1) \wedge t_1\neq t_2$.
\end{tabbing}
($\default\ F$ stands for $\caused\ \{F\}^{\rm ch}$ \cite{babb15action1}).

The fluents $\mi{Alive}$ and $\mi{Loaded}$ observe the commonsense law of inertia:
\begin{tabbing}
\ \ \ $\caused\ \{\mi{Alive}(t)\}^{\rm ch}\ \iif\ \top\ \after\ \mi{Alive}(t)$, \\
\ \ \ $\caused\ \{\sneg \mi{Alive}(t)\}^{\rm ch}\ \iif\ \top\ \after\ \sneg \mi{Alive}(t)$, \\
\ \ \ $\caused\ \{\mi{Loaded}\}^{\rm ch}\ \iif\ \top\ \after\ \mi{Loaded}$,\ \ \\
\ \ \ $\caused\ \{\sneg \mi{Loaded}\}^{\rm ch}\ \iif\ \top\ \after\ \sneg \mi{Loaded}$.
\end{tabbing}
We ensure no concurrent actions are allowed by stating
\begin{tabbing}
\ \ \ $\caused\ \bot\ \after\ a_1\wedge a_2$
\end{tabbing}
for every pair of action constants $a_1, a_2$ such that $a_1\neq a_2$.

Finally, we state that the initial values of all fluents are uniformly random ($b$ is a schematic variable that ranges over $\{\true, \false\}$):
\begin{tabbing}
\ \ \ $\caused\ \mi{Init\_Alive}(t)=\{\true: 0.5, \false: 0.5\}$,  \\
\ \ \ $\caused\ \mi{Init\_Loaded}=\{\true: 0.5, \false: 0.5\}$, \\
\ \ \ $\init\ \mi{Alive}(t)=b\ \iif\ \mi{Init\_Alive}(t)=b$,  \\
\ \ \ $\init\ \mi{Loaded}=b\ \iif\ \mi{Init\_Loaded}=b$.
\end{tabbing}

We translate the action description into an $\lpmln$ program and use {\sc lpmln2asp} to answer various queries about transition systems, such as prediction, postdiction and planning queries.\footnote{The complete {\sc lpmln2asp} program and the queries used in this section are given in \ref{ssec:yale-lpmln2asp}.}

\medskip\noindent
{\bf Prediction}\hspace{0.2cm} For a prediction query, we are given a sequence of actions and observations that occurred in the past, and we are interested in the probability of a certain proposition describing the result of the history, or the most probable result of the history. Formally, we are interested in the conditional probability
$$
Pr_{Tr(D, m)}(Result \mid Act, Obs)
$$
or
the MAP state
$$
  \underset{Result}{\rm argmax}Pr_{Tr(D, m)} (Result \mid Act, Obs)
$$
where $Result$ is a proposition describing a possible outcome, $Act$ is a set of facts of the form $i:a$ or $i:\sneg a$ for $a\in\sigma^{act}$, and $Obs$ is a set of facts of the form $i:c=v$ for $c\in\sigma^{fl}$ and $v\in Dom(c)$. 

In the Yale shooting example, such a query could be ``given that only the fat turkey is alive and the gun is loaded at the beginning, what is the probability that the fat turkey dies after shooting is executed?'' 
To answer this query, we manually translate the action description above into the input language of {\sc lpmln2asp} and add the following action and observation as constraints: 

\begin{lstlisting}
   :- not alive(slimTurkey, f, 0).
   :- not alive(fatTurkey, t, 0).
   :- not loaded(t, 0).
   :- not fire(fatTurkey, t, 0).
\end{lstlisting}
Executing the command
\begin{lstlisting}
   lpmln2asp -i yale-shooting.lpmln -q alive
\end{lstlisting}
yields
\begin{lstlisting}
   alive(fatTurkey, f, 1) 0.700000449318
\end{lstlisting}

\medskip\noindent
{\bf Postdiction}\hspace{0.2cm} In the case of postdiction, we infer a condition about the initial state given the history. Formally, we are interested in the conditional probability
$$
Pr_{Tr(D, m)}(Initial\_State \mid Act, Obs)
$$
or
the MAP state 
$$
\underset{Initial\_State}{\rm argmax}Pr_{Tr(D, m)}(Initial\_State \mid Act, Obs)
$$
where $Initial\_State$ is a proposition about the initial state; $Act$ and $Obs$ are defined as above.

In the Yale shooting example, such a query could be ``given that the slim turkey was alive and the gun was loaded at the beginning, the person shot at the slim turkey and it died, what is the probability that the fat turkey was alive at the beginning?''

Formalizing the query and executing the command
\begin{lstlisting}
   lpmln2asp -i yale-shooting.lpmln -q alive
\end{lstlisting}
yields
\begin{lstlisting}
   alive(fatTurkey, t, 0) 0.666661211973
\end{lstlisting}

\medskip\noindent
{\bf Planning}\hspace{0.2cm} In this case, we are interested in a sequence of actions that would result in the highest probability of a certain goal. Formally, we are interested in
{
\[
\underset{Act}{\rm argmax}\ Pr_{Tr(D, m)}(Goal\mid Initial\_State, Act)
\]
where $Goal$ is a {condition for a goal state}, and $Act$ is a sequence of actions $a\in\sigma^{act}$ specifying actions executed at each timestep.
}

In the Yale shooting example, such query can be ``given that both turkeys are alive and the gun is not loaded at the beginning, generate a plan that gives best chance to kill both the turkeys with 4 actions.''

Formalizing the query and executing the command

\BOCC
We describe the initial state and the goal as follows:
\begin{lstlisting}
:- not alive(``slimTurkey'', ``t'', 0).
:- not alive(``fatTurkey'', ``t'', 0).
:- not loaded(``f'', 0).

:- not alive(``slimTurkey'', ``f'', 4).
:- not alive(``fatTurkey'', ``f'', 4).
\end{lstlisting}
The command
\EOCC

\begin{lstlisting}
   lpmln2asp -i yale-shooting.lpmln
\end{lstlisting}
finds the most probable stable model, which yields
\begin{lstlisting}
   load(t,0)   fire(slimTurkey,t,1)   
   load(t,2)   fire(fatTurkey,t,3) 
\end{lstlisting}
which suggests to first kill the slim turkey and then the fat turkey.

\BOCC
\section{Parameter Learning in Probabilistic Action Domains}\label{sec:learning}

In a realistic setting, it is more practical to statistically derive the probability involved in an action domain from a collection of action and observation histories. We illustrate how this can be done with $\lpmln$ learning with the probabilistic variation of Yale shooting example discussed in Section \ref{sec:inference}.

As an example, suppose we have an action and observation history where the person attempted to kill the slim turkey three times by shooting at it, but only succeeded at the last attempt. The history can be represented as
\begin{lstlisting}
   :- not alive(``slimTurkey'', ``t'', 0).
   :- not alive(``fatTurkey'', ``t'', 0).
   :- not loaded(``t'', 0).
   :- not fire(``slimTurkey'', ``t'', 0).
   :- not alive(``slimTurkey'', ``t'', 1).
   :- not load(``t'', 1).
   :- not fire(``slimTurkey'', ``t'', 2).
   :- not alive(``slimTurkey'', ``t'', 3).
   :- not load(``t'', 3).
   :- not fire(``slimTurkey'', ``t'', 4).
   :- not alive(``slimTurkey'', ``f'', 5).
\end{lstlisting}
With the this history as the training data, we learn the weights of the following two $\lpmln$ rules:
\begin{lstlisting}
%%% caused Pf_Killed(SlimTurkey) = {t:?, f:?}
   @w(1) pf_turkeyKilled(``slimTurkey'', ``t'', ST).
   @w(2) pf_turkeyKilled(``slimTurkey'', ``f'', ST).
\end{lstlisting}
50 gradient ascent learning iterations yield
\begin{lstlisting}
   New weights:
   Rule 1:  -1.271
   Rule 2:  -0.729
\end{lstlisting}
The learned weights can be translated into probabilities. For example, the probability of successfully killing the turkey by shooting can be approximated by 
\[
\frac{exp(-1.271)}{exp(-1.271) + exp(-0.729)} = 0.368
\].
\EOCC

\section{Diagnosis in Probabilistic Action Domain}\label{sec:diagnosis}

One interesting type of reasoning tasks in action domains is diagnosis, where we observe a sequence of actions that fails to achieve some expected outcome and we would like to know possible explanations for the failure. Furthermore, in a probabilistic setting, we could also be interested in the probability of each possible explanation. In this section, we discuss how diagnosis can be automated in $p\cal{BC}$+ as probabilistic abduction and we illustrate the method through an example.
\subsection{Extending $p\cal{BC}$+ to Allow Diagnosis}

We define the following new constructs to allow probabilistic diagnosis in action domains. Note that these constructs are simply syntactic sugars that do not change the actual expressivity of the language.

\begin{itemize}
\item We introduce a subclass of regular fluent constants called {\em abnormal fluents}.
\item When the action domain contains at least one abnormal fluent, we introduce a special statically determined fluent constant $ab$ with the Boolean domain, and add 
$$
\default\ \sneg ab.
$$


\item We introduce the expression
\[
{\bf \causedab}\ F\ \iif\ G\ \after\ H
\]
where $F$ and $G$ are fluent formulas and $H$ is a formula, provided that $F$ does not contain statically determined constants and $H$ does not contain initpf constants. This expression is treated as an abbreviation of
\[
{\bf \caused}\ F\ \iif\ ab\wedge G\ \after\ H.
\]
\end{itemize}
Once we have defined abnormalities and how they affect the system, we can use
\[
\caused\ ab
\]
to enable taking abnormalities into account in reasoning.

\subsection{Example: Robot}
\label{ssec:robot}

The following example is modified from \cite{iwan02history}. Consider a robot located in a building with two rooms {\tt r1} and {\tt r2} and a book that can be picked up. The robot can move to rooms, pick up the book and put down the book. There is a $0.1$ chance that it fails when it tries to enter a room, a $0.2$ chance that the robot drops the book when it has the book, and a $0.3$ chance that the robot fails when it tries to pick up the book. The robot, as well as the book, was initially at {\tt r1}. It executed the following actions to deliver the book from {\tt r1} to {\tt r2}:
pick up the book; go to {\tt r2}; put down the book.
However, after the execution, it observes that the book is not at {\tt r2}. What is a possible reason?

We answer this query by modeling the action domain in the probabilistic action language as follows. We first introduce the following constants.
\vspace{0.2cm}
\hrule
\begin{tabbing}
Notation:  $r$ range over $\{\mi{R}_1, \mi{R}_2\}$. \\
Regular fluent constants:          \hskip 4cm  \=Domains:\\
$\;\;\;$ $\mi{LocRobot}$, $\;$ $\mi{LocBook}$   \>$\;\;\;$ $\{\mi{R}_1, \mi{R}_2\}$\\
$\;\;\;$ $\mi{HasBook}$                 \>$\;\;\;$ Boolean\\ 
Abnormal fluent constants:          \hskip 4cm  \=Domains:\\
$\;\;\;$ $\mi{EnterFailed}$, $\;$ $\mi{DropBook}$,  $\;$ $\mi{\mi{PickupFailed}}$                 \>$\;\;\;$  Boolean\\ 
Action constants:                          \>Domains:\\
$\;\;\;$ $\mi{Goto}(r)$, $\;$ $\mi{PickUpBook}$, $\;$ $\mi{PutdownBook}$                     \>$\;\;\;$ Boolean\\  
Pf constants:                          \>Domains:\\
$\;\;\;$ $\mi{Pf\_EnterFailed}$, $\;$ $\mi{Pf\_PickupFailed}$, $\;$ $\mi{Pf\_DropBook}$                     \>$\;\;\;$ Boolean\\ 
Initpf constants:                          \>Domains:\\
$\;\;\;$ $\mi{Init\_LocRobot}$, $\;$ $\mi{Init\_LocBook}$                     \>$\;\;\;$ $\{\mi{R}_1, \mi{R}_2\}$\\ 
$\;\;\;$ $\mi{Init\_HasBook}$                    \>$\;\;\;$ Boolean
\end{tabbing}
\hrule
\vspace{0.2cm}

The action $Goto(r)$ causes the location of the robot to be at $r$ unless the abnormality $\mi{EnterFailed}$ occurs:
\begin{tabbing}
\ \ \ $\caused\ \mi{LocRobot}\mvis r\ \after\ \mi{Goto}(r)\ \wedge \neg \mi{EnterFailed}$.
\end{tabbing}

Similarly, the following causal laws describe the effect of the actions $\mi{PickupBook}$ and $\mi{PutdownBook}$:
\begin{tabbing}
\ \ \ $\caused\ \mi{HasBook}\ \iif\ \mi{LocRobot}=\mi{LocBook}\ \after\ \mi{PickUpBook}\wedge \neg \mi{PickUpFailed}$ \\
\ \ \ $\caused\ \sneg\mi{HasBook}\ \after\ \mi{PutdownBook}$.
\end{tabbing}

If the robot has the book, then the book has the same location as the robot:
\begin{tabbing}
\ \ \ $\caused\ \mi{LocBook}=r\ \iif\ \mi{LocRobot}=r\land \mi{HasBook}$.
\end{tabbing}
The abnormality $\mi{DropBook}$ causes the robot to not have the book:
\begin{tabbing}
\ \ \ $\caused\ \sneg\mi{HasBook} \ \iif\ \mi{DropBook}$.
\end{tabbing}

The fluents $\mi{LocBook}$, $\mi{LocRobot}$ and $\mi{HasBook}$ observe the commonsense law of inertia:
\begin{tabbing}
\ \ \ $\caused\ \{\mi{LocBook}=r\}^{\rm ch}\ \after\ \mi{LocBook}=r$ \\ 
\ \ \ $\caused\ \{\mi{LocRobot}=r\}^{\rm ch}\ \after\ \mi{LocRobot}=r$\\
\ \ \ $\caused\ \{\mi{HasBook}=b\}^{\rm ch}\ \after\ \mi{HasBook}=b$.
\end{tabbing}

The abnormality $\mi{EnterFailed}$ has $0.1$ chance to occur when the action $Goto$ is executed:
\begin{tabbing}
\ \ \ $\caused\ \{\sneg\mi{EnterFailed}\}^{\rm ch}\ \iif \sneg\mi{EnterFailed}$\\
\ \ \ $\caused\ \mi{Pf\_EnterFailed}=\{\true:0.1, \false:0.9\}$\\ 
\ \ \ $\causedab\ \mi{EnterFailed}\ \iif\ \top\ \after\ \mi{pf\_EnterFailed}\wedge \mi{Goto}(r)$.
\end{tabbing}

Similarly, the following causal laws describe the condition and probabilities for the abnormalities $\mi{PickupFailed}$ and $\mi{DropBook}$ to occur:
\begin{tabbing}
\ \ \ $\caused\ \{\sneg\mi{PickupFailed}\}^{\rm ch}\ \iif \sneg\mi{PickupFailed}$\\ 
\ \ \ $\caused\ \mi{Pf\_PickupFailed}=\{\true:0.3, \false:0.7\}$\\ 
\ \ \ $\causedab\ \mi{PickupFailed}\ \iif\ \top\ \after\ \mi{Pf\_PickupFailed}\wedge \mi{PickupBook}$, \\ \\

\ \ \ $\caused\ \{\sneg\mi{DropBook}\}^{\rm ch}\ \iif \sneg\mi{DropBook}$\\ 
\ \ \ $\caused\ \mi{Pf\_DropBook}=\{\true:0.2, \false:0.8\}$\\ 
\ \ \ $\causedab\ \mi{DropBook}\ \iif\ \ \top\  \after\ \mi{Pf\_DropBook}\wedge \mi{HasBook}$.
\end{tabbing}
We ensure no concurrent actions are allowed by stating
\begin{tabbing} 
\ \ \ $\caused\ \bot\ \after\ a_1\wedge a_2$
\end{tabbing}
for every pair of action constants $a_1, a_2$ such that $a_1\neq a_2$.
Initially, it is uniformly random where the robot and the book is and whether the robot has the book:

\smallskip
\begin{tabbing}
\ \ \ $\caused\ \mi{Init\_LocRobot}=\{R_1:0.5, R_2:0.5\}$\\ 
\ \ \ $\caused\ \mi{Init\_LocBook}=\{R_1:0.5, R_2:0.5\}$\\ 
\ \ \ $\caused\ \mi{Init\_HasBook}=\{\true:0.5, \false:0.5\}$ \\
\ \ \ $\init\ \mi{LocRobot}=r\ \iif\ \mi{Init\_LocRobot}=r$\\
\ \ \ $\init\ \mi{LocBook}=r\ \iif\ \mi{Init\_LocBook}=r$\\
\ \ \ $\init\ \mi{HasBook}=b\ \iif\  \mi{Init\_HasBook}=b.$
\end{tabbing}

\smallskip\noindent
No abnormalities are possible at the initial state:
\begin{tabbing}
\ \ \ $\init\ \bot\ \iif\ \mi{EnterFailed}$\\ 
\ \ \ $\init\ \bot\ \iif\ \mi{PickupFailed}$ \\ 
\ \ \ $\init\ \bot\ \iif\ DropBook.$
\end{tabbing}


We add
$$
\caused\ ab
$$
to the action description to take abnormalities into account in reasoning and translate the action description into $\lpmln$ program, together with the actions that the robot has executed.\footnote{We refer the reader to \ref{ssec:robot-lpmln2asp} for the complete translation of the action description in the language of {\sc lpmln2asp}.}

Executing \lstinline|lpmln2asp -i robot.lpmln|
yields
\begin{lstlisting}
pickupBook(t,0)  ab(pickup_failed,t,1)  goto(r2,t,1) putdownBook(t,2)
\end{lstlisting}
which suggests that the robot fails at picking up the book.

\BOCC
By executing
\begin{lstlisting}
lpmln2asp -i robot.lpmln -q ab
\end{lstlisting}
we can see the probability of each abnormality:
\begin{lstlisting}
ab(pickup_failed, t, 1) 0.836065573764   ab(drop_book, t, 1) 0.0491803278688
ab(enter_failed, t, 2) 0.475409836063
\end{lstlisting}
\EOCC

Suppose that the robot has observed that the book was in its hand after it picked up the book. We expand the action history with
\begin{lstlisting}
:- not hasBook(t, 1).
\end{lstlisting} 
Now the most probable stable model becomes
\begin{lstlisting}
pickupBook(t,0)  goto(r2,t,1)  ab(drop_book,t,2)  putdownBook(t,2) 
\end{lstlisting} 
suggesting that robot accidentally dropped the book.

On the other hand, if the robot further observed that itself was not at {\tt r2} after the execution
\begin{lstlisting}
:- locRobot(r2, 3).
\end{lstlisting} 
Then the most probable stable model becomes
\begin{lstlisting}
pickupBook(t,0)  goto(r2,t,1)  ab(enter_failed,t,2)  putdownBook(t,2) 
\end{lstlisting} 
suggesting that the robot failed at entering {\tt r2}.


\BOCCC
In a realistic setting, it is not likely to have a theoretical probability for an abnormality to occur. It is more practical to statistically derive the probability from a collection of action and observation histories. We illustrate how this can be done with $\lpmln$ learning.

For this example, we provide a list of $12$ transitions as the training data. Each transition is a probabilistic stable model of $Tr(D, 1)$. In order to learn from multiple probabilistic stable models, we insert one more argument in every constant to represent the index of an instance (stable model). As an example, the first transition tells us that the robot performed {\tt goto} action to room {\tt r2}, which failed.
\begin{lstlisting}
:- not locRobot(r1'', 0, 1).   :- not locBook(``r2'', 0, 1).  :- not hasBook(``f'', 0, 1).
:- not goto(``r2'', ``t'', 0, 1).  :- not locRobot(``r1'', 1, 1).a
\end{lstlisting}
Among the training data, {\tt enter\_failed} occurred 1 time out of 4 attempts, {\tt pickup\_failed} occurred 2 times out of 4 attempts, and {\tt drop\_book} occurred 1 time out of 4 attempts. With the this training data, we learn the weights of the following $\lpmln$ rules:
\begin{lstlisting}
@w(1) pf_enterFailed(``t'',I) :- astep(I).   @w(2) pf_enterFailed(``f'',I) :- astep(I).
@w(3) pf_pickupFailed(``t'',I) :- astep(I).  @w(4) pf_pickupFailed(``f'',I) :- astep(I).
@w(5) pf_dropBook(``t'',I) :- astep(I).      @w(6) pf_dropBook(``f'',I) :- astep(I).
\end{lstlisting}
The learned weights can be translated into probabilities. For example, the probability of {\tt enter\_failed} can be computed as $\frac{exp(w(1))}{exp(w(1)) + exp(w(2))}$.
\EOCCC

\section{Related Work}

There exist various formalisms for reasoning in probabilistic action domains. $P{\cal C}$+ \cite{eiter03probabilistic} is a generalization of the action language $\cal{C}$+ that allows for expressing probabilistic information.  The syntax of $P\cal{C}+$ is similar to $p\cal{BC}$+, as both the languages are extensions of $\cal{C}+$. $P\cal{C}+$ expresses probabilistic transition of states through so-called {\em context variables}, which are similar to pf constants in $p\cal{BC}+$, in that they are both exogenous variables associated with predefined probability distributions. 
In $p\cal{BC}+$, in order to achieve meaningful probability computed through $\lpmln$, assumptions such as all actions have to be always executable and nondeterminism can only be caused by pf constants, have to be made. In contrast, $P\cal{C}+$ does not impose such semantic restrictions, and allows for expressing qualitative and quantitative uncertainty about actions by referring to the sequence of ``belief'' states---possible sets of states together with probabilistic information. 
On the other hand, the semantics is highly complex and there is no implementation of $P\cal{C}+$ as far as we know.

\cite{zhu12plog} defined a probabilistic action language called $\cal{NB}$, which is an extension of the (deterministic) action language $\cal{B}$. $\cal{NB}$ can be translated into P-log \cite{baral04probabilistic} and since there exists a system for computing P-log, reasoning in $\cal{NB}$ action descriptions can be automated. Like $p\cal{BC}+$ and $P\cal{C}+$, probabilistic transitions are expressed through dynamic causal laws with random variables associated with predefined probability distribution.
In $\cal{NB}$, however, these random variables are hidden from the action description and are only visible in the translated P-log representation. One difference between $\cal{NB}$ and $p\cal{BC}+$ is that in $\cal{NB}$ a dynamic causal law must be associated with an action and thus can only be used to represent probabilistic effect of actions, while in $p\cal{BC+}$, a fluent dynamic law can have no action constant occurring in it. This means state transition without actions or time step change cannot be expressed directly in $\cal{NB}$. Like p$\cal{BC}$+, in order to translate $\cal{NB}$ into executable low-level logic programming languages, some semantical assumptions have to be made in $\cal{NB}$. The assumptions made in $\cal{NB}$ are very similar to the ones made in $p\cal{BC}+$.

Probabilistic action domains, especially in terms of probabilistic effects of actions, can be formalized as Markov Decision Process (MDP). The language proposed in~\cite{baral02reasoning} aims at facilitating elaboration tolerant representations of MDPs. The syntax is similar to $p\cal{BC}+$. The semantics is more complex as it allows preconditions of actions and imposes less semantical assumption. The concept of {\em unknown variables} associated with probability distributions is similar to pf constants in our setting. There is, as far as we know, no implementation of the language. There is no discussion about probabilistic diagnosis in the context of the language. PPDDL \cite{littman04ppddl} is a probabilistic extension of the planning definition language PDDL. Like $\cal{NB}$, the nondeterminism that PPDDL considers is only the probabilistic effect of actions. The semantics of PDDL is defined in terms of MDP. There are also probabilistic extensions of the Event Calculus such as \cite{DAsaroBD017} and \cite{skarlatidis11probabilistic}.

In the above formalisms, the problem of probabilistic diagnosis is only discussed in \cite{zhu12plog}. \cite{balduccini03diagnostic} and \cite{baral00formulating}  studied the problem of diagnosis. However, they are focused on diagnosis in deterministic and static domains. \cite{iwan02history} has proposed a method for diagnosis in action domains with situation calculus. Again, the diagnosis considered there does not involve any probabilistic measure.

Compared to the formalisms mentioned here, the unique advantages of p$\cal{BC}$+ include its executability through $\lpmln$ systems, its support for probabilistic diagnosis, and the possibility of parameter learning in actions domains. 

$\lpmln$ is closely related to Markov Logic Networks \cite{richardson06markov}, a formalism originating from Statistical Relational Learning. However, Markov Logic Networks have not been applied to modeling dynamic domains due to its limited expressivity  from its logical part. 

\section{Conclusion}

$p{\cal BC}$+ is a simple extension of ${\cal BC}$+. The main idea is to assign a probability to each path of a transition system to distinguish the likelihood of the paths. The extension is a natural composition of the two ideas: In the semantics of ${\cal BC}$+, the paths are encoded as stable models of the logic program standing for the ${\cal BC}$+ description. Since $\lpmln$ is a probabilistic extension of ASP, it comes naturally that by lifting the translation to turn into $\lpmln$ we could achieve a probabilistic action language. 

In the examples above, the action descriptions, including the probabilities, are all hand-written. In practice, the exact values of some probabilities are hard to find. In particular, it is not likely to have a theoretical probability for an abnormality to occur. It is more practical to statistically derive the probability from a collection of action and observation  histories. For example, in the robot example in Section \ref{ssec:robot}, we can provide a list of action and observation histories, where different abnormalities occurred, as the training data. With this training data, we may learn the weights of the $\lpmln$ rules that control the probabilities of abnormalities.


\BOCC
\begin{lstlisting}
@w(1) pf_enterFailed(``t'',I) :- astep(I).   
@w(2) pf_enterFailed(``f'',I) :- astep(I).
@w(3) pf_pickupFailed(``t'',I) :- astep(I).  
@w(4) pf_pickupFailed(``f'',I) :- astep(I).
@w(5) pf_dropBook(``t'',I) :- astep(I).     
@w(6) pf_dropBook(``f'',I) :- astep(I).
\end{lstlisting}
The learned weights can be translated into probabilities. For example, the probability of
\[
{\tt enter\_failed}
\]
can be computed as 
\[
\frac{exp(w(1))}{exp(w(1)) + exp(w(2))}.
\]

{\cred
We are currently working on an $\lpmln$ learning system. The prototype system is available at \url{http://lpmln-learn.weebly.com}. 
{\cblu [[github or google code?]]}
The above task can be done with the system.}
\EOCC

Another future work is to build a compiler that automates the process of the translation of $p{\cal BC}$+ description into the input language of {\sc lpmln2asp}. 


\bigskip\noindent
{\bf Acknowledgements:} 
We are grateful to Zhun Yang and the anonymous referees for their useful comments. This work was partially supported by the National Science Foundation under Grant IIS-1526301.

\bibliographystyle{acmtrans}

\include{prob-action-tplp-appendix-0501-r}

\end{document}